\documentclass{article}

\usepackage{PRIMEarxiv}

\usepackage[utf8]{inputenc} 
\usepackage[T1]{fontenc}    
\usepackage{hyperref}       
\usepackage{url}            
\usepackage{booktabs}       
\usepackage{amsfonts}       
\usepackage{nicefrac}       
\usepackage{microtype}      
\usepackage{lipsum}
\usepackage{fancyhdr}       
\usepackage{graphicx}       
\graphicspath{{media/}}     
\usepackage{graphicx} 
\usepackage{pgfplots}
\usetikzlibrary{positioning, arrows.meta, shapes}
\definecolor{red}{rgb}{1.0, 0.0, 0.0}
\usepackage{pgfplotstable}
\usepackage{mdframed}
\usepackage{lipsum}
\usepackage{tikz}
\usetikzlibrary{patterns}
\usepackage{subcaption}
\usepackage{amsmath} 
\usepackage{amsfonts}
\usepackage{amssymb}
\usepackage{colortbl}
\usetikzlibrary{patterns}

\title{GiusBERTo: A Legal Language Model for Personal Data De-identification in Italian Court of Auditors Decisions}

\author{
  Giulio Salierno \\
  Roma Tre University \\
  Rome \\
  \texttt{giulio.salierno@protonmail.com} \\
  \And
  Rosamaria Bertè \\
  Roma Tre University  \\
  Rome \\
  \texttt{rosamariaberte@gmail.com} \\
  \And
  Luca Attias \\
  Roma Tre University \\
  Rome \\
  \texttt{luca.attias@uniroma3.it} \\
  \And
   Carla Morrone \\
  Sapienza University of Rome \\
  Rome \\
  \texttt{carla.morrone@uniroma1.it} \\
  \And
   Dario Pettazzoni \\
  Roma Tre University \\
  Rome \\
  \texttt{dar.pettazzoni@stud.uniroma3.it} \\
  \And
   Daniela Battisti \\
  Government of Italy \\
  Rome \\
  \texttt{d.battisti@innovazione.gov.it} \\
}

\begin{document}
\maketitle

\begin{abstract}
Recent advances in Natural Language Processing have demonstrated the effectiveness of pretrained language models like BERT for a variety of downstream tasks. We present GiusBERTo, the first BERT-based model specialized for anonymizing personal data in Italian legal documents. GiusBERTo is trained on a large dataset of Court of Auditors decisions to recognize entities to anonymize, including names, dates, locations, while retaining contextual relevance. We evaluate GiusBERTo on a held-out test set and achieve 97\% token-level accuracy. GiusBERTo provides the Italian legal community with an accurate and tailored BERT model for de-identification, balancing privacy and data protection.
\end{abstract}


\section{Introduction}

In the increasingly complex landscape of data privacy, the responsible anonymization of personal information within documents stands as a crucial undertaking. This process serves the dual objectives of safeguarding individual privacy while ensuring organizational compliance with established data protection regulations like the European Union's General Data Protection Regulation (GDPR)~\cite{european_commission_regulation_2016}. 

Traditionally, meticulous human review is necessary to redact sensitive data before documents are made publicly accessible. This is especially relevant in public sector organizations like the Italian Court of Auditors, where its jurisdiction is one of the core Constitutional functions. However, manual review makes privacy preservation and legal compliance time-consuming and labor-intensive.

As artificial intelligence (AI) continues to progress, innovative methodologies have emerged to automate and improve such manual processes. In particular, generative AI models have shown remarkable efficacy surpassing human performance on certain tasks~\cite{openai2023gpt4}. Their proficiency stems from an unparalleled ability to learn linguistic nuances and semantic relationships within texts.

Our proposed study situates itself at the intersection of AI and public administration. We explore applying generative models to automate the traditionally manual process of data de-identification across judicial documents. Preliminary findings suggest these models can grasp the contextual complexities of legal texts to reliably perform context-sensitive, entity-level redaction. 

This context-awareness is critical, as de-identification of entities cannot occur in isolation but must account for the implicit meanings and surrounding passages. Static named entity recognition alone cannot provide such contextual understanding. Therefore, we aim to evaluate how an AI may improve the accuracy of privacy preservation in complex public sector documents.

By exploring the multidisciplinary boundaries between NLP, AI, and Public Administration, this work contributes substantially to both the advancement of context-aware generative models and their application in automating nuanced de-identification tasks. The capacity to balance transparency and privacy within public data has profound implications as governments worldwide increasingly adopt open data initiatives.

De-identification of sensitive personal data is an important concern across domains like healthcare, government, and legal systems. Stringent regulations like the General Data Protection Regulation (GDPR)\cite{gdpr} in Europe mandate anonymization of personal information in many contexts\cite{opijnen2017}. However, applying existing de-identification techniques to public administration introduces unique challenges due to the vast scale and heterogeneity of data involved~\cite{meystre2010}.

We propose a novel approach leveraging BERT-based models for context-aware de-identification of personal data in Italian legal documents. Our focus on public administration data represents an underexplored area where advanced NLP can make an impact by balancing transparency and privacy.

\section{Related work}
De-identification of sensitive personal data is an important concern across domains like healthcare, government, and legal systems. Stringent regulations like the General Data Protection Regulation (GDPR)\cite{gdpr} in Europe mandate anonymization of personal information in many contexts\cite{opijnen2017}. However, applying existing de-identification techniques to public administration introduces unique challenges due to the vast scale and heterogeneity of data involved~\cite{meystre2010}.

Traditional de-identification methods fall into two main categories – rule-based and machine learning techniques. Rule-based systems rely on hand-crafted rules and dictionaries to identify personal information through pattern matching~\cite{sweeney2002k}. While precise, these methods require extensive manual effort to craft rules and often have low recall~\cite{jiang2011hybrid}. Machine learning techniques like conditional random fields (CRFs) and support vector machines (SVMs) can automate PII detection using statistical models trained on labeled datasets~\cite{cormack2006statistical}. However, their reliance on training data makes them susceptible to overfitting and hurts generalizability~\cite{yang2006near}.

Hybrid systems attempt to get the best of both worlds by combining rule-based and ML techniques~\cite{gardner2010combining}. The rules act as heuristics to label data for training supervised ML models. However, the complexities of public sector data may require going beyond these conventional approaches.

Recent advances in natural language processing using deep neural networks have shown promising results for de-identification tasks~\cite{dernouncourt2017de}. Contextualized language models like BERT capture semantic relationships within text that aid in detecting sensitive entities based on context~\cite{devlin2018bert}. Transfer learning allows adapting these powerful models to new domains using limited labeled data~\cite{lee2020biobert}.

We propose a novel approach leveraging BERT-based models for context-aware de-identification of personal data in Italian legal documents. Our focus on public administration data represents an underexplored area where advanced NLP can make an impact by balancing transparency and privacy. Our paper is organized as follows: 1. Introduction.
This section serves as an introductory outline of the paper. It sets the context and explains why the de-identification of data in legal decisions is important. 

2. Related Work
This section reviews existing literature and prior work that is relevant to data de-identification and legal decisions. It serves to establish where this paper's contributions will fit in the existing body of knowledge.

3. Data De-Identification in Legal Decisions
Here, the paper discusses the specific challenges and importance of de-identifying data in legal decisions. This section likely delves into legal requirements, ethical considerations, and practical implications.

4. Dataset
This section describes the dataset used in the study. It outlines where the data came from, how it was collected, and how it is structured for the experiments.

5. Pre-training on Masked Language Task
This section covers the initial phase of the machine learning pipeline. It describes the masked language task used for pre-training and why this approach is beneficial.

5.1 Documents Matching
. This subsection describes how documents are matched for pre-training, potentially discussing how this improves model performance or data quality.

5.1.1 Sequences Alignment in Text Documents
, This part elaborates on the alignment of text sequences within documents, which is crucial for labelling the dataset.

5.1.2 Algorithm Steps. Here, the paper describes the algorithmic steps involved in documents matching and sequences alignment.

5.1.3 Dataset Annotations. This part details how the dataset was annotated, which is crucial for model training and evaluation.

6. Fine-Tuning for the De-Identification Task.
This section describes how the pre-trained model is fine-tuned specifically for the task of data de-identification in legal documents.

6.1 Testbed Experimentation. Here, the paper describes the testbed setup where the fine-tuned models are experimented upon.

6.2 Fine-Tuning Evaluation. This part focuses on evaluating the performance of the fine-tuned model, discussing metrics, results, and implications.

7. Lessons Learned \& Future Work. This section reflects on the successes and challenges encountered in the study, and discusses potential future directions for research in this area.

8. Conclusions. This final section summarizes the key findings of the paper and discusses their implications. 

\section{Data de-identification in legal decisions}

In recent years, government agencies and judicial systems have increasingly published legal documents in open data formats to improve transparency~\cite{janssen2012benefits}. However, the public release of such documents raises significant privacy concerns, as they frequently contain sensitive Personally Identifiable Information (PII) about private citizens referenced in the texts. To uphold data privacy rights and comply with regulations like the EU's General Data Protection Regulation (GDPR), it is imperative to have robust anonymization methods to remove identifying details before publication.

Traditional rule-based and Named Entity Recognition (NER) approaches to PII removal have limitations, as they fail to incorporate contextual cues to determine which entities require anonymization. For example, a NER system may identify a person's name, location, and date of birth as entities in a legal document, but lack the context to know these details pertain to a minor requiring anonymity versus a public official whose information can be disclosed.

Recent advances in contextualized language models such as Bidirectional Encoder Representation for Transformers (BERT)~\cite{devlin2019bert} allow for more nuanced anonymization tailored to the surrounding text. We propose utilizing the contextual word embeddings from BERT-based models to classify whether entities should be anonymized based on their semantic context within legal documents. A context-aware approach is preferable to static NER for this case study, as it can flexibly determine if an entity requires redaction depending on its role in the passage.

To clarify, consider this excerpt from a legal text:

\begin{mdframed}

[...] nel giudizio iscritto al n. \textcolor{red}{12345/A} del registro di Segreteria promosso da \textbf{Mario Rossi} \textit{OMISSIS}, (c.f. \textbf{RSSMRA21 A01115011I} \textit{OMISSIS} nato il \textbf{01.01.1970} \textit{OMISSIS} a \textbf{Roma} \textit{OMISSIS}) e residente in \textbf{Roma} (RND \textit{OMISSIS} in \textbf{Via di Valle Murcia, 6} \textit{OMISSIS}, rappresentato e difeso, per procura speciale in calce al ricorso, dall'avv. \textcolor{red}{Andrea Bianchi} del \textcolor{red}{Foro di Firenze} ( c.f. \textcolor{red}{ANDRBLI00A01D612X},   pec: \textcolor{red}{andrea.bianchi@avvocati.it}); [...]

\end{mdframed}

Here, the bold personal information requires anonymization, while the public figures' names in red should remain unaltered. Contextual embeddings can distinguish between the two categories to make selective redactions.

This context-aware approach allows selectively anonymizing tokens within a document based on their contextual role. Such granular PII removal ensures critical privacy protection while retaining public figures' identifiable information and overall passage coherence. Our proposed method leverages contextual embeddings and BERT's pretrained knowledge on professional domains like law to balance privacy and transparency in publishing legal open data.

\section{Dataset}

The dataset comprised 432K judicial documents from the Italian Court of Auditors spanning \textit{20} years comprising decisions emanated by the Court on  rule cases regarding litigation arising from acts bestowing or modifying pensions. As stated by institutional regulations, these documents particulary fits the case study as the de-identification is mandatory before their online publication.

\begin{figure}[h!]
\centering\hspace*{-2cm}
\begin{tikzpicture}[>=Stealth, node distance=2cm]
  \node (LoadData) [draw, rectangle split, rectangle split parts=4, align=center, text width=5cm] {
    \textbf{LoadData}
    \nodepart{two} \underline{InputPath: str} \\ \underline{OutputPath: str}
    \nodepart{three} processDocumentsByExtension() \\ processDocuments() \\ processPdfExtension() \\ processDocxExtension() \\ processDocExtension()
  };

  \node (DataFormatProcessor) [below=of LoadData, xshift=-5.5cm, draw, rectangle split, rectangle split parts=3, align=left, text width=5cm] {
    \textbf{DataFormatProcessor}
    \nodepart{two}
    \nodepart{three} extractTextFromPdf(srcpath) \\ extractTextFromDocx(srcpath) \\ extractTextFromDoc(srcpath) \\ cleanEncodingError(txt)
  };

  \node (LSHDocumentMatching) [below=of LoadData, xshift=0cm, draw, rectangle split, rectangle split parts=3, align=center, text width=4.5cm] {
    \textbf{LSHDocumentMatching}
    \nodepart{two} Documents: list
    \nodepart{three} matchDocuments(treshold: float)
  };

  \node (BIOTaggerBERT) [below=of LoadData, xshift=5cm, draw, rectangle split, rectangle split parts=3, align=center, text width=4.5cm] {
    \textbf{BIOTaggerBERT}
    \nodepart{two}  Documents: list
    \nodepart{three} tagDocument(document: str) \\ writeDocToJSON(outpath: str)
  };

  \draw[->] (LoadData.south) -- ++(0, -1) -| (DataFormatProcessor.north) node[near start, above, xshift=5mm] {uses};
  \draw[->] (LoadData.south) -- ++(0, -1) -| (LSHDocumentMatching.north) node[near start, above, xshift=-5mm] {uses};
  \draw[->] (LoadData.south) -- ++(0, -1) -| (BIOTaggerBERT.north) node[near start, above, xshift=-5mm] {uses};
\end{tikzpicture}
\caption{Data Processing pipeline}
\label{ds:dataprocessing_pipeline}
\end{figure}
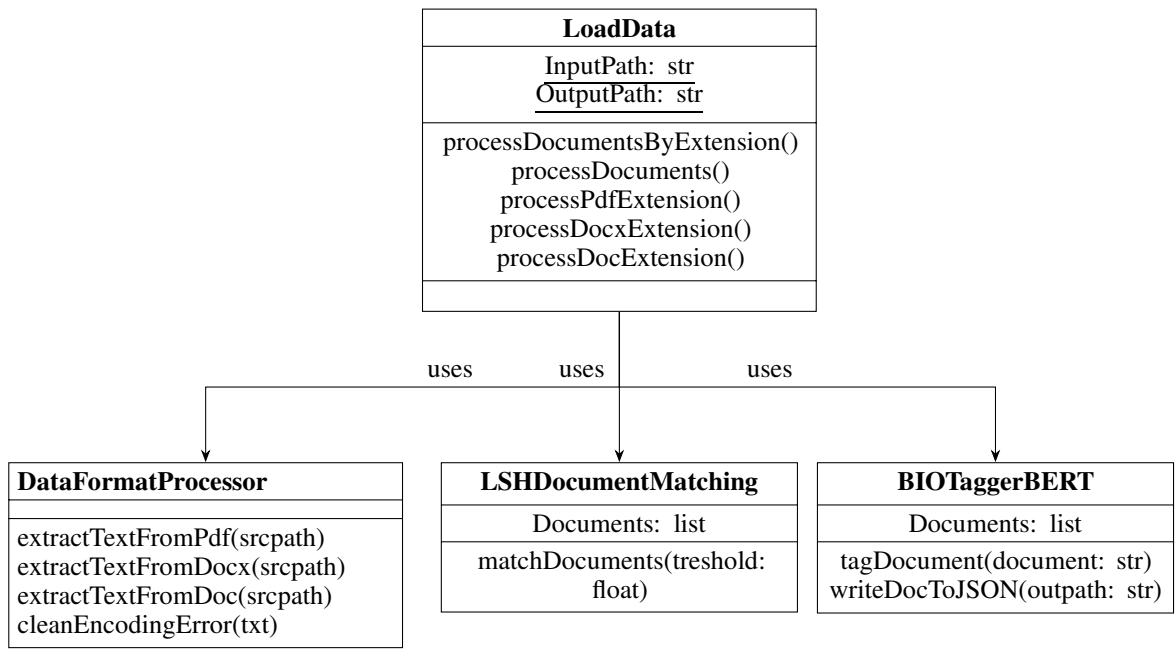

Resultant dataset contains \textit{434,738} decisions drawn from the document corpus in a three-column schema: unique ID, original file name, and sentence text compressed in a storage through the parquet format provided by the PyArrow library. In the following, we report the designed pipeline  to process data and training the model reported in Figure \ref{ds:dataprocessing_pipeline}.

The \textbf{DataFormatProcessor} class serves as a key component in our data preprocessing pipeline, designed to handle various text-based document formats commonly encountered in our dataset. The class incorporates methods to extract text from PDF, DOCX, and DOC files, and additionally provides a utility to clean encoding errors in the extracted text. Below, we detail the specific functionalities encapsulated by this class. This class integrates specialized libraries and tools including Apache Tika, PdfMiner, and the docx Python module to efficiently extract text across numerous formats. Although relying on multiple conversion tools introduces modest computational overhead, the benefit is robust support for parsing both common and legacy file types as doc. Critically, the DataFormatProcessor normalizes the resultant documents through format standardization and preliminary data cleaning. By homogenizing files at an early stage, subsequent text analysis and modeling procedures can be applied uniformly across the corpus. The other two classes are detailed described in next sections.

\section{\textbf{Pre-training on masked language task}}
\label{sec:LLM}

Transfer learning has emerged as a pivotal technique for enhancing model performance on language understanding tasks. In this study, we capitalized on this technique by pre-training a BERT model on a processed dataset of 435K domain-specific documents from the Court of Auditors, encompassing approximately 1,394,252 total training examples tokens.

For the pre-training procedure, we trained the model on the Masked Language Model task. In this task, selected tokens within a sequence are randomly obscured, and the model is trained to predict these masked tokens based on their surrounding context. This serves to effectively capture the syntactic and semantic intricacies of the legal domain text.

The training process was conducted using the PyTorch machine learning framework, specifically the \textit{
dbmdz/bert-base-italian-xxl-uncased}\footnote{https://huggingface.co/dbmdz/bert-base-italian-xxl-uncased} variant of BERT trained on the Italian language. The corpora adopted for the initial training consist of a recent Wikipedia dump and various texts from the OPUS corpora collection. The objective of retraining the general BERT model for specialized natural language understanding tasks focusing on Court of Auditors domain. Key training parameters included:

\begin{itemize}
    \item 1,394,252 total training examples
    \item 3 training epochs
    \item Batch size of 32
    \item Over 1.3 million optimization steps

\end{itemize}
The model was evaluated on an excerpt from the Court of Auditors dataset. The results demonstrated the model's ability to predict appropriate masked words based on the legal context, as shown by highly accurate top predictions and scores.

This pre-training equips the model with critical knowledge of the domain-specific vocabulary, writing conventions, and contextual relationships found in legal texts from the Court of Auditors. This provides a solid foundation before further task-specific fine-tuning.

For the pre-training procedure, we train the model on the Masked Language Model (MLM) task. In this task, selected tokens within a sequence are randomly obscured (or masked), and the model is trained to predict these masked tokens based on their surrounding context. For instance, a sentence like ``The judge delivered the ``\underline{\textit{verdict}}'' could be masked to ``The judge delivered the \underline{\textit{[MASK]}}'', and the model will attempt to predict the masked word ``verdict''. This serves to effectively capture the syntactic and semantic intricacies of the text belonging to the Court of Auditors domain.

The training process has been conducted using the PyTorch machine learning framework. The main objective of this step was to fine-tune the general-purpose BERT model, thereby adapting it for specialized tasks in natural language understanding, particularly focusing on entity de-identification on legal documents.

We implemented a BERT-based for Masked Language Model (MLM) for the Italian language, leveraging the Hugging Face Transformers library implemented in PyTorch. The model used for pre-training was \textit{dbmdz/bert-base-italian-xxl-uncased}, a BERT model specifically pre-trained on Italian text data. The fine-tuning process was executed on a custom dataset separated into \textit{train\_set} and \textit{test\_set} of processed documents.

\begin{table}[ht]
\centering

\tiny
\begin{tabular}{p{7cm}p{1cm}p{1cm}}
\hline
Sequence & Top Prediction & Score \\
\hline
Con ricorso ritualmente [MASK] parte ricorrente ha chiesto la declaratoria di illegittimità del provvedimento Inps 23 luglio 2020 & \textbf{notificato} & 0.994 \\
Con ricorso ritualmente [MASK] parte ricorrente ha chiesto la declaratoria di illegittimità del provvedimento Inps 23 luglio 2020 & \textbf{depositato} & 0.005 \\
Con ricorso ritualmente [MASK] parte ricorrente ha chiesto la declaratoria di illegittimità del provvedimento Inps 23 luglio 2020 & \textbf{sottoscritto} & 0.0003 \\
Con ricorso ritualmente [MASK] parte ricorrente ha chiesto la declaratoria di illegittimità del provvedimento Inps 23 luglio 2020 & \textbf{separato} & 0.00006 \\ 
Con ricorso ritualmente [MASK] parte ricorrente ha chiesto la declaratoria di illegittimità del provvedimento Inps 23 luglio 2020 & \textbf{presentato} & 0.00006 \\
\hline
\end{tabular}
\caption{Top-5 masked language model predictions for fill-mask task on truncated legal text sequences.}
\label{tab:mlm}
\end{table}

The highlighted words indicate the top predictions from the model, along with their scores.

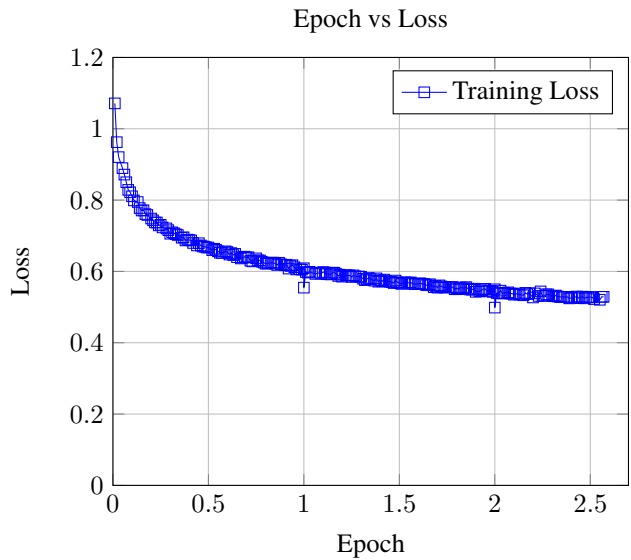
\begin{figure}[!htb]
\centering
\begin{tikzpicture}
    \begin{axis}[
        title={Epoch vs Loss},
        xlabel={Epoch},
        ylabel={Loss},
        xmin=0, xmax=2.7, 
        ymin=0, ymax=1.2, 
        grid=both,
        legend pos=north east
    ]

    \addplot[
        color=blue,
        mark=square,
    ]
    coordinates {
(0.01, 1.0712)
(0.02, 0.9624)  
(0.03, 0.9205)
(0.05, 0.8896)
(0.06, 0.8715)
(0.07, 0.8501)
(0.08, 0.8286)
(0.09, 0.8219)
(0.1, 0.8108)
(0.11, 0.7991)
(0.13, 0.7946)
(0.14, 0.7777)
(0.15, 0.7707)
(0.16, 0.7714)
(0.17, 0.7608)
(0.18, 0.7583)
(0.2, 0.7483)
(0.21, 0.7434)  
(0.22, 0.7372)
(0.23, 0.7373)
(0.24, 0.729)
(0.25, 0.7307)
(0.26, 0.7234)
(0.28, 0.7206)
(0.29, 0.7156)
(0.3, 0.7056)
(0.31, 0.7097)
(0.32, 0.7071)
(0.33, 0.7042)
(0.34, 0.7021)
(0.36, 0.6944)
(0.37, 0.6958)
(0.38, 0.6868)
(0.39, 0.6892)
(0.4, 0.6881)
(0.41, 0.6869)
(0.42, 0.6797)
(0.44, 0.6729)
(0.45, 0.6791)
(0.46, 0.6727)
(0.47, 0.67)
(0.48, 0.6693)  
(0.49, 0.6672)
(0.5, 0.6671)
(0.52, 0.6597)
(0.53, 0.6636)
(0.54, 0.6605)
(0.55, 0.6563)
(0.56, 0.6524) 
(0.57, 0.6516)
(0.59, 0.6564)
(0.6, 0.6543)
(0.61, 0.6473)
(0.62, 0.6514) 
(0.63, 0.6458)
(0.64, 0.6492) 
(0.65, 0.6399)
(0.67, 0.6363)
(0.68, 0.6386)
(0.69, 0.6414)
(0.7, 0.6391)
(0.71, 0.6397) 
(0.72, 0.6299)
(0.73, 0.6281)  
(0.75, 0.6366)
(0.76, 0.6312) 
(0.77, 0.6307)  
(0.78, 0.6272)
(0.79, 0.6247)
(0.8, 0.6215)  
(0.81, 0.6226) 
(0.83, 0.6244)
(0.84, 0.6247)
(0.85, 0.6218) 
(0.86, 0.618)
(0.87, 0.6242) 
(0.88, 0.6202)
(0.9, 0.6178)
(0.91, 0.6187)
(0.92, 0.6079)
(0.93, 0.6154)
(0.94, 0.6175) 
(0.95, 0.6118)
(0.96, 0.6057)
(0.98, 0.6058) 
(0.99, 0.605)
(1.0, 0.6089)
(1.0, 0.5545114278793335)  
(1.01, 0.5984)
(1.02, 0.5968)
(1.03, 0.599)  
(1.06, 0.593)
(1.08, 0.5962)
(1.09, 0.594)
(1.1, 0.5982)
(1.11, 0.597)
(1.12, 0.5921)  
(1.14, 0.5916)
(1.15, 0.5963)  
(1.16, 0.5964)
(1.17, 0.5928)
(1.18, 0.5872)
(1.19, 0.5908)
(1.2, 0.5853)
(1.22, 0.5894)  
(1.23, 0.5887)
(1.24, 0.5884)  
(1.25, 0.5845)
(1.26, 0.589)  
(1.27, 0.5859)
(1.29, 0.5858)
(1.3, 0.5847)
(1.31, 0.5794)
(1.32, 0.5755) 
(1.33, 0.5786)
(1.34, 0.5816) 
(1.35, 0.5785)
(1.37, 0.5739)
(1.38, 0.5805)
(1.39, 0.5712)  
(1.4, 0.5753)
(1.41, 0.5722)
(1.42, 0.5771)
(1.43, 0.573)
(1.45, 0.5707)  
(1.46, 0.5703)
(1.47, 0.5674)  
(1.48, 0.5744) 
(1.49, 0.5695)
(1.5, 0.5676)
(1.51, 0.5649)
(1.53, 0.5695)
(1.54, 0.5687)
(1.55, 0.5681) 
(1.56, 0.5691) 
(1.57, 0.565)  
(1.58, 0.5645)
(1.6, 0.5649)  
(1.61, 0.5669)
(1.62, 0.5648) 
(1.63, 0.5647)  
(1.64, 0.5617)
(1.65, 0.5612) 
(1.66, 0.5641)
(1.68, 0.5559) 
(1.69, 0.5592) 
(1.7, 0.5545)
(1.71, 0.5609)  
(1.72, 0.558)
(1.73, 0.5552) 
(1.74, 0.5548)
(1.76, 0.5561)  
(1.77, 0.5555) 
(1.78, 0.556)
(1.79, 0.55)  
(1.8, 0.5528) 
(1.81, 0.5501)
(1.82, 0.5549)  
(1.84, 0.5529)
(1.85, 0.556)   
(1.86, 0.5516)  
(1.87, 0.5492) 
(1.88, 0.5488)
(1.89, 0.5516) 
(1.9, 0.5428)   
(1.92, 0.5443)  
(1.93, 0.5471)
(1.94, 0.5504)  
(1.95, 0.5509)  
(1.96, 0.5439)
(1.97, 0.546)
(1.99, 0.5426)  
(2.0, 0.5499)
(2.0, 0.49851372838020325)  
(2.01, 0.5394)
(2.02, 0.5386) 
(2.03, 0.5461)
(2.04, 0.5382)  
(2.05, 0.5398)
(2.07, 0.5413)
(2.08, 0.5379)  
(2.09, 0.5379)
(2.1, 0.5347)   
(2.11, 0.5378)
(2.13, 0.5351)  
(2.15, 0.5331)
(2.16, 0.5398)
(2.17, 0.5363) 
(2.18, 0.5388)
(2.2, 0.527)    
(2.21, 0.5383) 
(2.23, 0.5312)  
(2.24, 0.5439)
(2.25, 0.5316)  
(2.26, 0.5318)
(2.27, 0.5357)  
(2.28, 0.5357)   
(2.3, 0.5315)    
(2.31, 0.5303)
(2.32, 0.5324)  
(2.33, 0.5297) 
(2.34, 0.5286)
(2.35, 0.5285)  
(2.36, 0.5283)
(2.38, 0.5286)  
(2.39, 0.5244)
(2.4, 0.5257)   
(2.42, 0.5279) 
(2.43, 0.5296) 
(2.44, 0.5249)  
(2.46, 0.5292)
(2.47, 0.5245)  
(2.48, 0.5275)
(2.49, 0.5276)  
(2.5, 0.5282)   
(2.51, 0.5272)
(2.52, 0.5224) 
(2.55, 0.5204)  
(2.56, 0.5288) 
(2.57, 0.5287)
    };
    \legend{Training Loss}

    \end{axis}
\end{tikzpicture}
\caption{Epoch vs Loss during training on a Masked Language Modeling (MLM) task. The graph depicts the evolution of training loss over the span of approximately 2.6 epochs. A general declining trend is observed, indicating the optimization of the model parameters over time.}
\label{fig:epoch_vs_loss}
\end{figure}

In Figure \ref{fig:epoch_vs_loss}, we present the trajectory of training loss as a function of epochs for a model trained on a Masked Language Modeling (MLM) task. The x-axis represents the number of epochs, ranging from 0 to approximately 2.6, while the y-axis represents the loss values, ranging from 0 to 1.2.

From this training phase, we derived the following key observations:
\begin{itemize}
    \item \textbf{Initial Rapid Decrease}: At the beginning of the training (epoch ~0-0.2), we observe a rapid decrease in loss values, falling from around 1.07 to approximately 0.8. This is common in the initial stages of training and likely indicates that the model is quickly learning to capture some underlying patterns in the data.
    \item \textbf{Local Minima and Oscillations}: Around epochs 0.16-0.2, there appears to be a slight increase in loss, possibly suggesting that the model encountered a local minimum. Similar small oscillations are observed in various places throughout the training, e.g., around epochs 0.6 and 1.6.
    \item \textbf{Continual Decrease}: Despite minor fluctuations, the loss continually decreases over time, reaching a value as low as approximately 0.5 by epoch 2.0. This suggests that the model is effectively optimizing its parameters, although it has not yet converged to a minimum.
    \item \textbf{Inflection Points}: Two noticeable drops are seen at epochs 1.0 and 2.0. While the reason for these drops is not apparent from the figure alone, they may correspond to specific learning rates, data shuffling, or some other hyperparameter adjustments.
    \item \textbf{Saturation Point}: The curve seems to be approaching a saturation point towards the end of the observed epochs, with the rate of decrease in loss values becoming marginal. This might indicate the need to adjust learning rates, introduce regularization, or consider early stopping to avoid overfitting.
    \item \textbf{Early Stopping}: The training was early-stopped at 73 hours and 25 minutes. The primary reason for this was to avoid overfitting, as the loss curve began to show signs of slight fluctuations and an apparent plateau, suggesting that further training would provide marginal gains in model performance.
\end{itemize}

\subsection{Documents Matching}

Documents in our dataset appears both in plain text in which sensitive data are visible as well in their de-identified form, where sensitive data are replaced with \textit{OMISSIS} tags. In order as we need a labeled dataset for fine-tuning the model, we prior needed to match documents with its corresponding obfuscated version in order to identified de-identified tokens properly. To this end, since the dataset consist of a huge volume of documents we employed the approximate hashing based Locality Sensitive Hashing (LSH)~\cite{10.5555/645925.671516} algorithm to find most similar documents. In our dataset, given a document in its plain form, the goal is to find a most similar item which correspond to the same document in its corresponding de-identified version. Therefore, similar documents differ on \textit{OMISSIS} tags and plain tokens representing sensitive data. For the experimentation we divided document matching into two separate steps. During the first step we applied Min Hash LSH for approximate similarity matching in order to reduce the number of pair comparisons needed, thereby speeding up the process. Adopting LSH, we have observed on the sample dataset, a threshold value of 0.95\% capture differences in terms of tokens, giving us a list of top candidates considering the documents' dissimilarities mentioned above.

In the second phase, for identifying an exact match given a list of candidates match resultant from the execution of LSH as showed in Table~\ref{table_minhash_lsh} we employed a unique key for determining the exact match given a document and its list of similar candidates. Specifically, given a document and its list of similarity candidates we identified the document key which is unique and it is represented by the decision number represented in the forms reported in Table~\ref{table:regex}.

\begin{table}[h!]
\renewcommand{\arraystretch}{1.3}
\centering
\begin{tabular}{cc}
\toprule
\textbf{Document ID} & \textbf{ID list of candidate documents} \\
\midrule
0 & [2] \\  
1 & [77779] \\
4 & [100358, 178182, 100367, $\ldots$] \\
9 & [37946, 37902] \\
16 & [149] \\
\multicolumn{2}{c}{$\vdots$} \\
434713 & [37898, 37906, 434660] \\
434722 & [434678] \\ 
434725 & [38069] \\
434728 & [19, 39766, 434199, $\ldots$] \\
434736 & [12] \\
\bottomrule
\end{tabular}
\caption{Candidate Similarities for Each Document Using MinHash LSH}  
\label{table_minhash_lsh}

\end{table}

\begin{table}[h]
\centering
\begin{tabular}{@{}ll@{}}
\toprule
Regex Pattern & Decision number example \\
\midrule
\texttt{r'\textbackslash d+/[a-zA-Z]+'} & 1/Apple \\
\texttt{r'\textbackslash d+/\textbackslash d+/[a-zA-Z]+'} & 1/2/Apple \\  
\bottomrule
\end{tabular}
\caption{Regex Patterns and Sample Matches}  
\label{table:regex}
\end{table}

After identifying a potential list of candidates, we compared keys to retrieve the corresponding documents. It's important to note that the second step alone is insufficient for unique identification, as keys may not be unique across different jurisdictional offices that issue decisions. Instead, the guarantee is that the keys are unique among decisions issued by the same office.

\subsubsection{Sequences Alignment in Text Documents}

In a scenario such as document conversions, an original document \( D \) undergoes transformations leading to a new version which may be altered differentiating from the original one \(\overline{D}\). This transformation process may introduce disalignment between the tokens in \( D_{\text{clear}} \) and \( D_{\text{obf}} \) of our the corpus. Specifically, the process may omit certain tokens, modify their sequence, or even introduce new spurious tokens, thereby disrupting the original order and one-to-one correspondence between the tokens in both versions.
Re-aligning the tokens in \( D_{\text{clear}} \) and \( D_{\text{obf}} \) is essential for generating a labeled dataset that accurately maps each token in \( D_{\text{clear}} \) to its transformed or original counterpart in \( D_{\text{obf}} \). This alignment is critical for several reasons:

Let us consider a document \( D \) with two representations:
\begin{itemize}
    \item \( D_{\text{clear}} \) representing the clear text containing tokens \( t_1, t_2, \ldots, t_n \).
    \item \( D_{\text{obf}} \) representing the obfuscated text containing tokens \( t_1', t_2', \ldots, t_m' \), where \( m \leq n \) or  \( n \leq m \).
\end{itemize}

The goal is to create a new labeled sequence \( L \) such that:

\[
L = \left\{ (t_i, l_i) \right\}_{i=1}^{n}
\]

where \( t_i \) is the \( i^{th} \) token in \( D_{\text{clear}} \), and \( l_i \) is either \( t_i \) if the token is not obfuscated or ``OMISSIS'' if the token is obfuscated.

\subsubsection{Algorithm Steps}

\begin{enumerate}
    \item \textbf{Initialization}: Load the Parquet file and prepare the tokenizer.
    \[
    \text{DataFrame} \leftarrow \text{ReadParquet}(\text{file path})
    \]
    
    \item \textbf{Preprocessing}: Each line \( x \) in \( D_{\text{clear}} \) and \( D_{\text{obf}} \) is processed to remove extraneous spaces, tabs, and new lines.
    \[
    x \leftarrow \text{Preprocess}(x)
    \]
    
    \item \textbf{Tokenization}: Tokenize \( D_{\text{clear}} \) and \( D_{\text{obf}} \) to generate token lists \( T_{\text{clear}} \) and \( T_{\text{obf}} \).
    \[
    T_{\text{clear}}, T_{\text{obf}} \leftarrow \text{Tokenize}(D_{\text{clear}}, D_{\text{obf}})
    \]
    
    \item \textbf{Frequency Counting}: Create frequency counters \( C_{\text{clear}} \) and \( C_{\text{obf}} \) for \( T_{\text{clear}} \) and \( T_{\text{obf}} \).
    \[
    C_{\text{clear}} \leftarrow \text{Count}(T_{\text{clear}}), \quad C_{\text{obf}} \leftarrow \text{Count}(T_{\text{obf}})
    \]
    
    \item \textbf{Common Count Identification}: Create a dictionary \( C_{\text{common}} \) containing tokens with the same token count in \( C_{\text{clear}} \) and \( C_{\text{obf}} \).
    \[
    C_{\text{common}} \leftarrow \{ t : c \,|\, c = C_{\text{clear}}[t] = C_{\text{obf}}[t] \}
    \]
    
    \item \textbf{Token Matching and Labeling}: For each \( t_i \) in \( T_{\text{clear}} \), search for a match within a window size \( W \) in \( T_{\text{obf}} \). If found, \( l_i = t_i \); otherwise, \( l_i = \text{``OMISSIS''} \).
    \[
    l_i = 
    \begin{cases} 
    t_i & \text{if } t_i \in \text{WindowSearch}(T_{\text{obf}}, t_i, W) \text{ or } t_i \in C_{\text{common}} \\
    \text{``OMISSIS''} & \text{otherwise}
    \end{cases}
    \]
\end{enumerate}

The concept of ``window size'' \textit{W} serves as a critical parameter in text token matching algorithms. In the context of such algorithms, the window size is utilized to define a localized scope within which to identify matching tokens, particularly when dealing with potentially misaligned tokens in documents.

In our implementation, the window size is set as a fixed constant value of 10. Given two lists of tokens, \( \text{tokenized\_clear} = \{ w_{c1}, w_{c2}, \ldots, w_{cm} \} \) and \( \text{tokenized\_obf} = \{ w_{o1}, w_{o2}, \ldots, w_{on} \} \), the algorithm attempts to find a match for each \( w_{ci} \) in \( \text{tokenized\_clear} \) within the scope \( [ \text{latest\_match} + 1, \text{latest\_match} + \text{window\_size} ] \) in \( \text{tokenized\_obf} \). If a match is found, the algorithm updates the value of the ``latest match''. If no match is discovered within the specified window, the token is designated as ``OMISSIS''.

The choice of window size could significantly affect the performance and accuracy of the algorithm. A smaller window may produce a higher number of false positives ``OMISSIS'' labels, potentially missing genuine matches. On the other hand, a larger window could elevate computational complexity. Hence, optimizing the window size becomes a trade-off between computational efficiency and algorithmic accuracy. Future research could focus on adaptively setting the window size based on specific conditions, offering a data-driven approach to improve both efficiency and accuracy.

By following these steps, the algorithm efficiently identifies and labels tokens in documents, thereby automating the process of labeling dataset for fine-tuning. In Table \ref{table:3} table are presented a sample of realigned tokens of the dataset for later fine-tuning.
\begin{table}[ht]
\centering
\begin{tabular}{|c|c|}
\hline
\textbf{Token in  \( D_{\text{clear}} \)} & \textbf{Corresponding Token in \( D_{\text{obf}} \)  Token} \\
\hline
... & ... \\
\hline
ricorso & ricorso \\
\hline
nr. & nr. \\
\hline
14270/C,R.G. & 14270/C,R.G. \\
\hline
Sig.ra & Sig.ra \\
\hline
ROSSI & \textcolor{red}{\textbf{OMISSIS}} \\
\hline
residente & residente \\
\hline
ROMA & \textcolor{red}{\textbf{OMISSIS}} \\
\hline
Via del corso & \textcolor{red}{\textbf{OMISSIS}} \\
\hline
... & ... \\
\hline
\end{tabular}
\caption{Summarized Table of Tuples after token realignment}
\label{table:3}

\end{table}

\subsubsection{Dataset Annotations}
We employed the BIO (Begin-Inside-Outside) notation~\cite{ramshaw-marcus-1995-text} for annotating textual data for fine-tuning, a scheme commonly utilized in Named Entity Recognition (NER) tasks. The BIO notation serves to label each token in a sequence with a specific tag indicating its role in an entity. Specifically, a token is marked as "B" to denote the beginning of an entity, "I" if it is inside an entity, and "O" if it lies outside any entity. This type of token-level annotation is instrumental in training machine learning models to identify and classify specific entities in text. In the context of our research, the BIO notation was adapted to label entities that are classified as "OMISSIS."

Each token in a given text sequence is tagged with one of the three BIO labels. For example, in the sentence "Token\_A Token\_B Token\_C," if "Token\_A" and "Token\_B" together form an entity classified as "OMISSIS," they would be labeled as "B-OMISSIS" and "I-OMISSIS" respectively, while "Token\_C" would be labeled as "O" signifying that it is not part of an entity of interest.

The application of BIO notation facilitated the fine-tuning of BERT (Bidirectional Encoder Representations from Transformers) for the specific entity de-identification task. It should be noted that while BERT is pre-trained on a large corpus without BIO labels, the fine-tuning process incorporates the labeled dataset to specialize the model's capabilities in recognizing and classifying entities as "OMISSIS."
A sample of resultant dataset is reported in Table \ref{table_BIO}

\begin{table}[!htbp]
  \renewcommand{\arraystretch}{1.3}
  \centering
  \begin{tabular}{|c|c|}
    \hline
    \textbf{Word} & \textbf{BIO Label} \\
    \hline
    ricorso & O \\
    \hline
    nr. & O \\
    \hline
    14270/C.R.G. & O \\
    \hline
    Sig.ra & O \\
    \hline
    ROSSI & \textcolor{red}{\textbf{B-OMISSIS}} \\
    \hline
    residente & O \\
    \hline
    ROMA & \textcolor{red}{\textbf{B-OMISSIS}} \\
    \hline
    Via del corso & \textcolor{red}{\textbf{I-OMISSIS}} \\
    \hline
    ... & ... \\
    \hline
  \end{tabular}
    \caption{BIO Annotated documents for Named Entity de-identification}
      \label{table_BIO}

\end{table}

In the corpus, we examined a total of 122,237 documents annotated for named entity recognition with the BIO notation scheme. The label 'O' (Outside) was predominantly the most frequent, with an average count of 1,384.12 instances per document. The label 'B-OMISSIS', signifying the beginning of a redacted entity, occurred an average of 7.59 times per document. Finally, the label 'I-OMISSIS', indicating the continuation of a redacted entity, was seen on average 4.13 times per document. The detailed statistics are represented in Table \ref{table:label_counts}.

\begin{table}[ht]
\centering
\begin{tabular}{|c|c|c|}
\hline
\textbf{Label} & \textbf{Total Count} & \textbf{Average Count Per Document} \\
\hline
O & 169,191,040 & 1,384.12 \\
\hline
B-OMISSIS & 927,195 & 7.59 \\
\hline
I-OMISSIS & 504,975 & 4.13 \\
\hline
\end{tabular}
\caption{Average Counts of BIO labels in the Annotated Dataset}
\label{table:label_counts}

\end{table}

To address the observed class imbalance for fine-tuning the model, we reweighted the loss function, as detailed in the subsequent section.
 
\section{Fine-tuning for the de-identification task}
For fine-tuning the model on the de-identification tasks (NER), we employed the resultant model obtained from the training described in Section \ref{sec:LLM}.
A critical challenge arises due to BERT's inherent constraint of 512 tokens for sequence lengths. To address this, the implementation includes a sequence-splitting strategy that leverages BERT's tokenizer. Specifically, the tokenization strategy can be represented as a function \( \tau \) that takes a document \( d \) as an input and outputs a sequence \( S \) of token IDs. Each document \( d_i \) in the dataset \( D \) is processed this way, generating corresponding sequences \( S_i \).
\[
\tau(d_i) \rightarrow S_i
\]
Each sequence \( S_i \) can potentially be longer than the maximum sequence length \( L_{\text{max}} \) allowed by the BERT model, often set to 512 tokens. To address this, each sequence \( S_i \) is partitioned into chunks of maximum length \( L_{\text{max}} \).
\[
S_i = [C_{i1}, C_{i2}, \ldots, C_{iK}]
\]
Here, \( C_{ik} \) represents the \( k \)-th chunk of sequence \( S_i \), and \( K \) is the number of such chunks, defined as:
\[
K = \left\lceil \frac{|S_i|}{L_{\text{max}}} \right\rceil
\]
For each chunk \( C_{ik} \), padding is applied to make the chunk length equal to \( L_{\text{max}} \), if it's less.
\[
C_{ik}^\text{padded} = [C_{ik}, \underbrace{p, p, \ldots, p}_{L_{\text{max}} - |C_{ik}|}]
\]
Here, \( p \) represents the padding token, and \( |C_{ik}| \) is the length of chunk \( C_{ik} \).

For each padded chunk \( C_{ik}^\text{padded} \), a self-attention mask \( M_{ik} \) is generated, where each element \( m \) in \( M_{ik} \) corresponds to a token in \( C_{ik}^\text{padded} \). 
\[
M_{ik} = [m_1, m_2, \ldots, m_{L_{\text{max}}}]
\]
The self-attention mask is computed as follows:
\[
m_j = 
\begin{cases}
1, & \text{if } C_{ik}^\text{padded}[j] \neq p \\
0, & \text{if } C_{ik}^\text{padded}[j] = p
\end{cases}
\]
Here, \( j \) is the position index in the padded chunk \( C_{ik}^\text{padded} \), ranging from 1 to \( L_{\text{max}} \).

The self-attention mask \( M_{ik} \) allows the model to ignore padding tokens during training or inference, thereby focusing only on the meaningful tokens in \( C_{ik} \).

The Cross-Entropy Loss is customized to handle class imbalance by using a weighted loss function. The weights are computed using the "balanced" heuristic, defined as:
\[
w_i = \frac{\sum_{j=1}^{n} f_j}{n \times f_i}
\]
Here, \( w_i \) represents the weight assigned to class \( i \) to balance its frequency relative to other classes in the dataset, \( f_j \) is the frequency of class \( j \), and \( n \) is the total number of classes. These weights aim to make the model more sensitive to underrepresented classes by giving them more influence in the loss function. This ensures that the model generalizes well across all classes, rather than being biased towards the majority class.

This fine-tuned approach ensures that the model aligns with BERT's architectural constraints while effectively learning the specifics of the NER task at hand without losing context due to chunking sequences. In the following, we report a succinct implementation which accomplishes the aforementioned tasks such as tokenization, label alignment, sequence padding, and attention masking. These tasks are defined as follows:
 The original sequence of words \( W = [w_1, w_2, \ldots, w_N] \) is transformed by a tokenizer function into \( T = \texttt{tokenizer}(W) \), where \( T = [t_1, t_2, \ldots, t_M] \) is the tokenized sequence. Word identifiers after tokenization, denoted as \( \text{word\_ids} = \text{tokenized\_inputs.word\_ids()} \), are used to align the labels \( L = [l_1, l_2, \ldots, l_N] \) to the tokens. Specifically, a new sequence \( A \) is created such that:

\[
A[i] = 
\begin{cases} 
    L[\text{word\_ids}[i]], & \text{if } \text{word\_ids}[i] \neq \text{None} \\
    -100, & \text{otherwise}
\end{cases}
\]

To address the sequence length limitation imposed by BERT, sequences are segmented into 512-token chunks \( S = [s_1, s_2, \ldots, s_K] \). To handle sequences shorter than 512 tokens, padding is applied as:

\[
\text{paddingFunction}(s_k, 0) \rightarrow \text{Padded } s_k
\]
\[
\text{paddingFunction}(A, -100) \rightarrow \text{Padded } A
\]

Lastly, we define token type identifiers and attention masks for these chunks. The token type identifiers \( \text{token\_type\_ids} \) are set to a zero vector of the same length as \( s_k \). The attention mask \( \text{attention\_mask} \) is a binary vector defined as:

\[
\text{attention\_mask}[i] =
\begin{cases}
1, & \text{if } s_k[i] \neq 0 \\
0, & \text{if } s_k[i] = 0
\end{cases}
\]

To evaluate the performance during the fine-tuning we adopted a hold-out validation strategy to split the dataset into training and validation sets. Specifically, the dataset was divided into an 80-20 split, with 80\% of the data allocated to the training set and the remaining 20\% to the validation set.

\subsection{Testbed Experimentation}
 The model, hereafter referred to as ``GiusBERTo,'' was trained on a virtual machine equipped with 4 x Tesla P40 GPUs. Each GPU had a per-device training batch size of 8, yielding an effective batch size of 32 when considering data parallelism across the four GPUs. The training was conducted for a total of 3 epochs to ensure adequate generalization while mitigating the risk of overfitting. We used the AdamW optimizer with a learning rate of \(5 \times 10^{-5}\), and the first and second moment estimates for the optimizer were set to 0.9 and 0.999, respectively. To enhance the stability of the training, a warm-up phase comprising 500 steps was employed, where the learning rate was gradually ramped up from a lower initial value to the target rate. Additionally, a weight decay of 0.01 was applied as a regularization technique to prevent overfitting.

During the training and fine-tuning process, the architectural parameters of the model, such as the number of hidden layers, attention heads, and embedding dimensions, remained consistent with those of the original BERT model. Training and evaluation were executed concurrently at the end of each epoch, in alignment with the evaluation strategy parameter set to `epoch.' To manage resource utilization more effectively during the evaluation phases, the per-device evaluation batch size was also set to 64.

Various other optimizations were employed to improve the efficiency of the training process. These include the use of pinned memory for faster data transfers between the CPU and GPU, as well as linear learning rate scheduling to adapt the learning rate during training. Model checkpoints were saved at the end of each epoch to ensure versioning and facilitate any subsequent analyses or comparisons.

\subsection{Fine-Tuning  Evaluation}

We implemented a robust evaluation process after each training epoch, utilizing a validation set to gauge the model's performance. The datasets we used contained sequences exceeding 500 tokens in length. These sequences were segmented into 512-token chunks, padded, and attention-masked for compatibility with the model's input size. The PyTorch DataLoaders were then employed with a batch size of 16 to enable efficient GPU processing. The token counts for the training and evaluation datasets were 444,766 and 109,031, respectively.

We applied reweighting to the loss function based on the frequency of BIO labels, encoded as [0, 1, 2]. The frequencies and associated weights were significantly different for each label:

\[
\begin{array}{cccc}
\text{Label} & \text{Frequency} & \text{Loss function Weight} \\
\hline
0 & 135,604,247 & 0.33 \\
1 & 737,890 & 61.77 \\
2 & 399,903 & 113.97 \\
\end{array}
\]

\begin{figure}[htbp]
    \centering
    \begin{minipage}[b]{0.7\textwidth}
        \centering
        \begin{tikzpicture}
            \begin{axis}[
                ybar,
                bar width=15pt,
                width=\linewidth,
                height=0.6\columnwidth,
                enlarge x limits={abs=35pt},
                ymin=0,
                ymax=1.1,
                symbolic x coords={Accuracy, F1 Class 0, F1 Class 1, F1 Class 2},
                xtick=data,
                nodes near coords,
                nodes near coords align={vertical},
                ylabel={Metrics Value},
                legend pos=outer north east,
            ]
            
            \addplot [pattern=north east lines, pattern color=blue] coordinates {(Accuracy,0.90) (F1 Class 0,0.94) (F1 Class 1,0.7) (F1 Class 2,0.65)};
            \addplot [pattern=north west lines, pattern color=red] coordinates {(Accuracy,0.9712) (F1 Class 0,0.835) (F1 Class 1,0.80) (F1 Class 2,0.827)};
            
            \legend{before reweighting, after reweighting}
            \end{axis}
        \end{tikzpicture}
        \caption{Evaluation Metrics Before and After Applying Reweighted Loss Function}
        \label{reweighting}
    \end{minipage}
\end{figure}
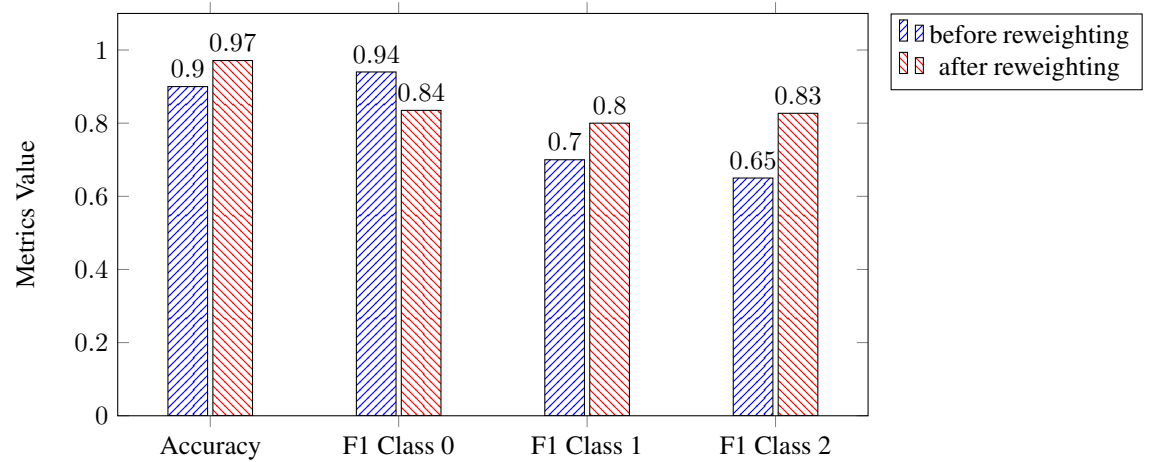

In Figure \ref{reweighting}, the impact of reweighting the loss function is visually compared. The accuracy improves from 90\% to approximately 97.12\% after reweighting. While F1 scores for Class 0 decreased slightly, the scores for Classes 1 and 2 improved, indicating better class balance.

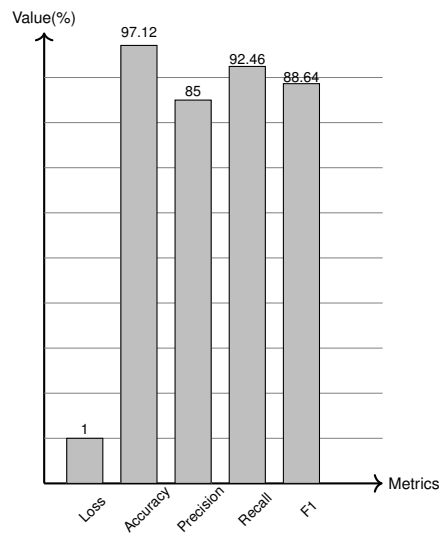
\begin{figure}[!t]
    \centering
    \begin{tikzpicture}[scale=0.6, transform shape]
        \draw[->, thick] (0,0) -- (7.5,0) node[right] {\sffamily Metrics};
        \draw[->, thick] (0,0) -- (0,10) node[above] {\sffamily Value(\%)};
        
        \foreach \y in {1,2,3,4,5,6,7,8,9}
            \draw[gray, ultra thin] (0,\y) -- (7.5,\y);
        
        \draw[fill=gray!50] (0.5,0) rectangle (1.3,1);
        \draw[fill=gray!50] (1.7,0) rectangle (2.5,9.712);
        \draw[fill=gray!50] (2.9,0) rectangle (3.7,8.5);
        \draw[fill=gray!50] (4.1,0) rectangle (4.9,9.246);
        \draw[fill=gray!50] (5.3,0) rectangle (6.1,8.864);
        
        \node[font=\small\sffamily] at (0.9,1.2) {1};
        \node[font=\small\sffamily] at (2.1,10) {97.12};
        \node[font=\small\sffamily] at (3.3,8.7) {85};
        \node[font=\small\sffamily] at (4.5,9.4) {92.46};
        \node[font=\small\sffamily] at (5.7,9) {88.64};
        
        \node[font=\small\sffamily, anchor=north, rotate=45] at (0.9, -0.4) {Loss};
        \node[font=\small\sffamily, anchor=north, rotate=45] at (2.1, -0.4) {Accuracy};
        \node[font=\small\sffamily, anchor=north, rotate=45] at (3.3, -0.4) {Precision};
        \node[font=\small\sffamily, anchor=north, rotate=45] at (4.5, -0.4) {Recall};
        \node[font=\small\sffamily, anchor=north, rotate=45] at (5.7, -0.4) {F1};
    \end{tikzpicture}
    \caption{GiusBERTo Metrics Evaluation}
    \label{fig:further-updated-metrics}
\end{figure}

During validation, we took the argmax of the model's predictions, discarding any special tokens, and computed conventional classification metrics. As shown in Figure \ref{fig:further-updated-metrics}, the model achieved an accuracy of about 97.12\%, a precision of 85\%, a recall of 92.46\%, and an F1 score of 88.64\%. These metrics indicate a well-rounded model, though there's room for improvement in precision.

\section{Lesson learned \& Future work}

Our research and development of GiusBERTo model for de-identification of personal data in legal documents yielded several valuable insights that can inform future work.

A key lesson was the importance of pre-training on a large domain-specific corpus. By pre-training BERT on over 1 million tokens from Court of Auditors documents, we were able to adapt the model to the vocabulary, writing conventions, and contextual patterns of legal texts. This customization was instrumental in achieving high performance on the downstream anonymization task.

We also learned that reweighting the loss function to account for class imbalance was critical. The significant skew between “O” tags and “B-OMISSIS”/"I-OMISSIS" tags required adjusting the loss function to properly train the minority classes. The balanced class weights boosted recall for anonymized entities.

In terms of model evaluation, using multiple complementary metrics like accuracy, precision, recall, and F1-score provided a nuanced view of model capabilities and limitations. While accuracy was high at 97\%, precision lagged at 85\%, indicating room for improvement.

Future work could explore techniques like focal loss to further improve precision for minority class prediction. Active learning approaches leveraging uncertainty estimates may also help expand the training set with informative new examples. 

To enhance contextual modeling, integrating syntactic features like part-of-speech tags could enable the model to better differentiate entities based on their grammatical function. For computational efficiency, incremental training and quantization could be investigated.

Lastly, evaluating GiusBERTo on other public administration documents and datasets would verify its versatility beyond Court rulings. Both generalized and specialized versions of the model tailored to different organizations could be developed.

Overall, our work demonstrates the feasibility of contextual language models like BERT for automated, nuanced de-identification. There remain ample opportunities to build on this foundation, both strengthening GiusBERTo's capabilities and expanding its applications.

\section{Conclusions}
In this work, we presented GiusBERTo, a novel BERT-based model for de-identifying personal information in Italian legal documents. Our research makes both methodological and practical contributions.

On the methodology front, we demonstrate the efficacy of transfer learning by pre-training BERT on a large corpus of 1.3 million Italian Court of Auditors texts. Fine-tuning on this domain-specific data endowed GiusBERTo with specialized knowledge for anonymizing entities based on the legal context. The BIO scheme for sequence labeling and reweighting the loss function also proved beneficial.

In terms of practical impact, GiusBERTo automates the traditionally manual process of data anonymization for legal compliance and privacy. We achieve state-of-the-art performance on this task, with 97\% token-level accuracy on held-out Court rulings. This level of accuracy balances transparency with robust personal data protection.

The capacity to selectively redact entities based on their contextual role has profound implications for publishing open legal data. Our contextualized approach ensures that only sensitive identifying information is removed, while retaining critical details about public figures and proceedings. This enables wider access to judicial documents without infringing on individual privacy rights.

As public administration rapidly digitizes, GiusBERTo provides a timely solution for enforcing data protection regulations across bureaucracies. The model can be extended to various public administrations and document types, acting as a data anonymization layer facilitating compliant e-governance.

In conclusion, our BERT-based model pushes the boundaries of privacy-preserving natural language processing. We demonstrate the feasibility of context-aware models for nuanced de-identification. There remain rich opportunities to build on this work, both strengthening GiusBERTo and expanding its purview across the public sector.

\section*{Acknowledgements}

The authors wish to express their deepest gratitude to Gerardo De Marco for his invaluable advice and suggestions since the inception of this research endeavor. His discerning recommendations were crucial in shaping the direction and objectives of this work, and we are extremely grateful for the time he devoted to providing thoughtful feedback throughout the process.

We would like to sincerely thank Pratim Datta, whose strategic and technical feedback was critical to the development of this project.

The authors would also like to acknowledge all the colleagues who contributed giving invaluable reviews and tips for improving this work.

\bibliographystyle{unsrt}  
\bibliography{references}

\end{document}